\documentclass[letterpaper, 10 pt, conference]{ieeeconf}  

\IEEEoverridecommandlockouts                              

\usepackage{cite}
\usepackage{amsmath,amssymb,amsfonts}
\usepackage{algorithmic}
\usepackage{graphicx}
\usepackage{textcomp}
\usepackage{xcolor}
\usepackage{graphicx}
\usepackage{multirow}
\usepackage{adjustbox}
\usepackage{booktabs}
\usepackage{microtype}
\usepackage{threeparttable}
\usepackage{siunitx}
\usepackage[hang,flushmargin]{footmisc}
\usepackage[breaklinks,colorlinks]{hyperref}

\def\BibTeX{{\rm B\kern-.05em{\sc i\kern-.025em b}\kern-.08em
    T\kern-.1667em\lower.7ex\hbox{E}\kern-.125emX}}
\begin{document}

\title{\LARGE \bf
Evidential Uncertainty Estimation for Multi-Modal Trajectory Prediction}

\author{Sajad Marvi$^{1,2}$, Christoph Rist$^{1,3}$, Julian Schmidt$^{1}$, Julian Jordan$^{1}$, and Abhinav Valada$^2$
\thanks{$^{1}$Mercedes-Benz AG, Germany.}%
\thanks{$^{2}$Department of Computer Science, University of Freiburg, Germany.}%
\thanks{$^{3}$Intelligent Vehicles Group, TU Delft, The Netherlands.}%
}

\maketitle
\thispagestyle{empty}
\pagestyle{empty}

\begin{abstract}
Accurate trajectory prediction is crucial for autonomous driving, yet uncertainty in agent behavior and perception noise makes it inherently challenging. While multi-modal trajectory prediction models generate multiple plausible future paths with associated probabilities, effectively quantifying uncertainty remains an open problem. In this work, we propose a novel multi-modal trajectory prediction approach based on evidential deep learning that estimates both positional and mode probability uncertainty in real time. Our approach leverages a Normal Inverse Gamma distribution for positional uncertainty and a Dirichlet distribution for mode uncertainty. Unlike sampling-based methods, it infers both types of uncertainty in a single forward pass, significantly improving efficiency. Additionally, we experimented with uncertainty-driven importance sampling to improve training efficiency by prioritizing underrepresented high-uncertainty samples over redundant ones. We perform extensive evaluations of our method on the Argoverse~1 and Argoverse~2 datasets, demonstrating that it provides reliable uncertainty estimates while maintaining high trajectory prediction accuracy.
\end{abstract}

\section{Introduction}
Autonomous driving relies on accurate trajectory prediction to anticipate the future motion of agents and ensure safe navigation. However, due to the inherent uncertainty in human driving behavior, environmental dynamics, and perception errors, deterministic trajectory prediction often fails to capture the full range of possible outcomes. To mitigate this problem, recent deep learning models employ multi-modal trajectory prediction, where multiple plausible future trajectories are generated along with their likelihoods. Despite this progress, a key challenge remains: how to quantify and leverage uncertainty in real-time to improve trajectory prediction model robustness and training efficiency.

\begin{figure}
\centering
\includegraphics[width=\linewidth]{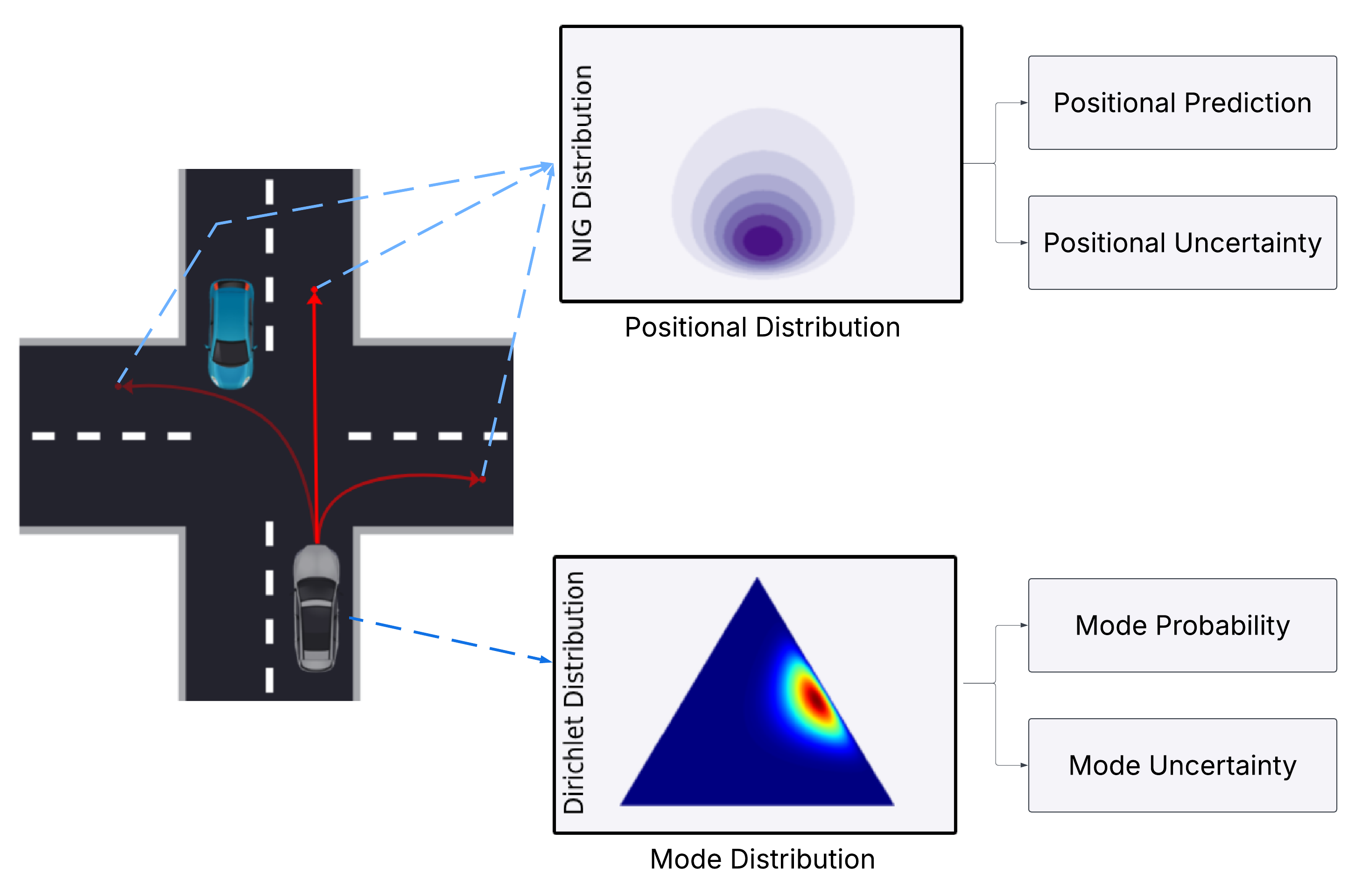}
\caption{Evidential priors are used for both positional and mode probability prediction. At each time step, a Normal Inverse Gamma distribution models a higher-order probability distribution over the mean (\(\mu\)) and variance (\(\sigma^2\)) of each predicted position component (\(x\) or \(y\)) which means each sample from NIG distribution would be a Gaussian distribution over the (\(x\) or \(y\)). For modes probability prediction, a Dirichlet distribution estimates the probabilities of different possible trajectories, quantifying uncertainty in decision-making.}
\label{fig:intro}
\vspace{-0.3cm}
\end{figure}

Uncertainty in trajectory prediction arises due to various factors, including limited training data, measurement noise, and model complexity \cite{sensoy2018evidential}. Hence, estimating the uncertainty is crucial for having a reliable autonomous agent, as high-uncertainty predictions often indicate rare, complex, or ambiguous driving scenarios that require additional attention. To address this problem, we leverage the deep evidential framework~\cite{aminiEDL,sensoy2018evidential} to model uncertainty at multiple levels, from predicted position uncertainty at each timestep to trajectory and scene-level uncertainty. Specifically, we model position coordinates using a \textit{Normal Inverse Gamma (NIG)} prior, enabling the model to predict not only the most probable future positions but also their associated uncertainty. Additionally, we employ a \textit{Dirichlet prior} over trajectory likelihoods, treating the probability assignment to each trajectory as a classification problem (Fig.~\ref{fig:intro}). Traditional sampling-based methods are computationally expensive and time-consuming due to the need for multiple forward passes during inference to estimate uncertainty. However, our evidential-based approach provides real-time uncertainty estimation with significantly reduced inference times. Furthermore, compared to one-shot uncertainty estimators that may lack the robustness of sampling-based methods, our approach strikes a balance by providing efficient and reliable uncertainty quantification.


Beyond improving uncertainty estimation, we explore its role in enhancing training efficiency. Large-scale datasets, such as Argoverse~2~\cite{wilsonArgoverse}, predominantly consist of driving straight lane scenarios that are relatively easy for the model to learn. However, the most valuable learning opportunities arise from complex interactions and rare maneuvers, which significantly contribute to improving model performance. At the same time, leading companies in autonomous driving are generating enormous amounts of data to train state-of-the-art models, creating challenges in terms of data storage, processing, and efficiency. Storing and utilizing all the collected data is becoming increasingly impractical. Therefore, a more strategic approach is necessary. Since the evidential framework is designed to predict higher uncertainty for unseen or underrepresented cases, we leverage this property to augment the training set selectively with high-uncertainty samples. By prioritizing these informative scenarios, we reduce the need for storing and processing vast amounts of redundant data while ensuring that the model focuses on learning from the most meaningful and diverse cases. This strategy enhances the model’s robustness and enables it to generalize effectively to challenging, real-world conditions.

To summarize, our main contributions are:
\begin{itemize}
    \item We introduce a \textit{real-time uncertainty estimation} framework based on the deep evidential framework, eliminating the need for expensive sampling-based approaches.
    \item We model uncertainty at multiple levels, using a \textit{Normal Inverse Gamma (NIG) prior} for positional uncertainty and a \textit{Dirichlet prior} for mode probability uncertainty.
    \item We propose an \textit{importance sampling strategy} that uses predicted uncertainty to improve training efficiency, reducing dataset size while maintaining performance.
    \item We benchmarked our method against widely used uncertainty estimation techniques and demonstrated its superior performance.
\end{itemize}

\section{Related Work}

{\parskip=0pt
\noindent\textit{Trajectory Prediction Models}: Trajectory prediction has been a significant focus in autonomous driving and multi-agent systems, where the task is to predict the future trajectory of agents (e.g., vehicles, pedestrians) based on their past behavior and the surrounding environment. Early models relied on simple kinematic models or heuristic approaches, such as constant velocity models or polynomial regression~\cite{xie2017vehicle}, which assumed minimal external influence on vehicle motion. With advancements in deep learning, trajectory prediction models have evolved beyond rule-based methods to data-driven approaches. Recurrent Neural Networks~(RNNs)~\cite{graves2013generating}, Long Short-Term Memory (LSTM) networks~\cite{zyner2017long}, and Temporal Convolution Networks (TCN)~\cite{radwan2020multimodal} have been widely adopted due to their ability to model sequential dependencies and learn complex temporal relationships. However, they often struggle with the multi-modal nature of future behaviors. More recent works have introduced Graph Neural Networks (GNNs) to model spatial relationships between agents in the environment, capturing interactions through graph structures~\cite{buchner20223d}. At the same time, Transformer-based methods~\cite{huang2022multi, ngiam2021scene} have gained popularity for capturing long-range dependencies in multi-agent environments. The combination of GNNs for spatial relationships and Transformers for temporal dependencies has led to state-of-the-art performance in multi-agent motion forecasting~\cite{zhou2022hivt, zhou2023query}.

{\parskip=2pt
\noindent\textit{Uncertainty Estimation Models}: Uncertainty estimation plays a critical role in trajectory prediction, particularly in applications such as autonomous driving, where understanding the reliability of predictions is crucial for safe decision-making. Traditional trajectory prediction models often produce deterministic outputs, but these outputs may not fully capture the uncertainty inherent in predicting future behavior. To address this, uncertainty estimation techniques are commonly applied to trajectory prediction models. Based on the taxonomy presented in~\cite{he2025surveyuncertaintyquantificationmethods}, uncertainty quantification for DNNs can be divided into two categories: \textit{Epistemic Uncertainty} (Model Uncertainty) and \textit{Aleatoric Uncertainty} (Data Uncertainty). Epistemic uncertainty arises from a lack of knowledge due to insufficient training data, suboptimal model architecture, or distribution shifts. It is reducible with better data or improved models. Methods such as Bayesian Neural Networks (BNNs)~\cite{blundell2015weight,gal2016dropout}, ensemble models~\cite{lakshminarayanan2017simple,mallick2022deep}, and distance-aware neural networks~\cite{liu2020simple} aim to capture epistemic uncertainty. Aleatoric uncertainty stems from inherent randomness in data, such as sensor noise or ambiguous labels, and is irreducible. It can be modeled using deep discriminative models~\cite{kendall2017uncertainties} or deep generative models~\cite{shu2024zero}.}

Currently, few methods claim to capture both types of uncertainty effectively. \textit{Evidential Deep Learning} is one such approach, providing an efficient alternative to sampling-based Bayesian methods. Unlike Monte Carlo (MC) dropout or deep ensembles, which require multiple forward passes during inference, evidential methods estimate uncertainty in a single pass, significantly improving computational efficiency. However, while the classification variant of evidential deep learning has been successfully applied to various perception tasks~\cite{sirohi2023uncertainty,sirohi2023uncertaintylidar,nallapareddy2023evcenternet,mohan2024panoptic}, its regression counterpart has received limited attention in real-world applications. Conformal Prediction~\cite{karimi2023quantifying} is an alternate approach that provides a measure of total uncertainty but does not distinguish between epistemic and aleatoric uncertainty explicitly. However, recent studies suggest a strong correlation between the aleatoric and epistemic uncertainty predicted by models, indicating that methods aiming to disentangle these uncertainties may still capture overlapping information~\cite{mucsanyi2024benchmarking}.

\section{Method}
In this section, we first define the problem that we are addressing and then present our proposed approach that provides an effective solution.

\begin{figure*}
\centering
\includegraphics[width=\textwidth]{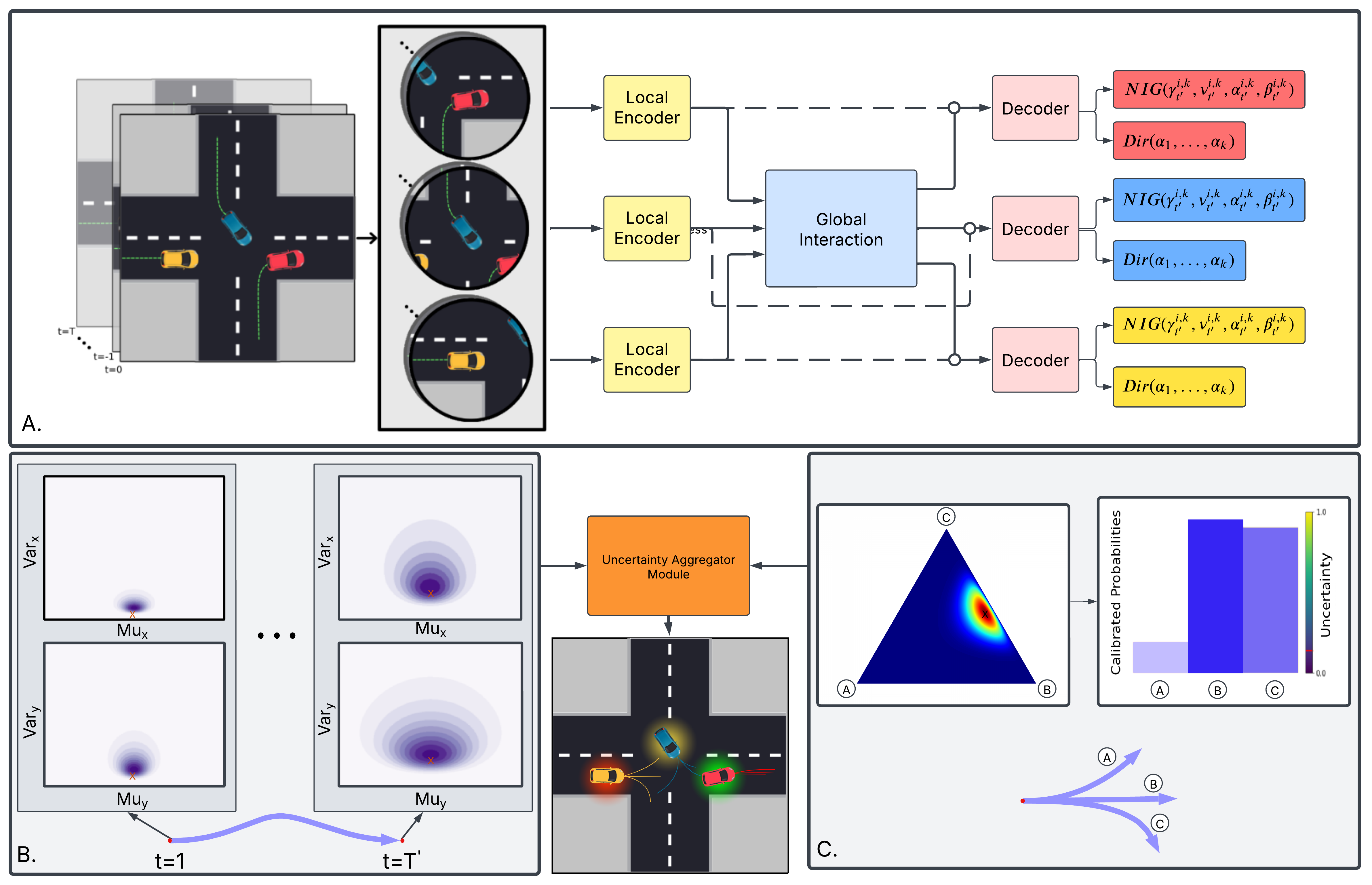}
\caption{Overview of our evidential-based multi-modal trajectory prediction framework. (A) The trajectory prediction model consists of three stages: Local Context Extraction, Global Intercation Modeling, and Multi-modal future decoding, which predict the parameters distribution. (B) The evidential framework models position uncertainty using a NIG distribution, capturing a higher-order distribution over the mean and variance of future positions. This allows real-time estimation of both aleatoric and epistemic uncertainty at each timestep. (C) A Dirichlet distribution models the categorical probability distribution over trajectory modes. Finally, both positional and mode uncertainties are processed by the Uncertainty Aggregator module to compute a holistic uncertainty estimate for each agent.}
\label{fig:evidential}
\vspace{-0.3cm}
\end{figure*}

\subsection{Problem Formulation}

Let \(\mathbf{X}_{1:T}^i = \{(x_t^i, y_t^i)\}_{t=1}^T\) denote the observed trajectory of an agent \(i\) in a scene, where \((x_t^i, y_t^i)\) are the coordinates of the agent in a 2D plane at time \(t\). The goal is to predict \(K\) future potential trajectories for the agent on a prediction horizon \(T'\), represented as \(\mathbf{\hat{Y}}_{1:T'}^{i,k} = \{(\hat{x}_{t'}^{i,k}, \hat{y}_{t'}^{i,k})\}_{t'=1}^{T'}\), where \(k \in \{1, 2, \dots, K\}\) denotes the mode index. Each mode corresponds to a distinct plausible future trajectory.
To account for uncertainty in the predictions, each predicted mode \(k\) is associated with a predictive uncertainty \(\mathbf{\Sigma}_{1:T'}^{i,k} = \{(\sigma_{x,t'}^{i,k}, \sigma_{y,t'}^{i,k})\}_{t'=1}^{T'}\), where \(\sigma_{x,t'}^{i,k}\) and \(\sigma_{y,t'}^{i,k}\) represent the uncertainties in the \(x\) and \(y\) coordinates at time \(t'\), respectively. Furthermore, each mode is associated with a mode probability \(p_k^i \in [0, 1]\), which indicates the likelihood of mode \(k\) being the correct representation of the agent's future trajectory, with the constraint that the sum of the probabilities over all modes is equal to one: \(\sum_{k=1}^K p_k^i = 1\).

The input to the prediction model includes:
\begin{enumerate}
    \item \textit{Agent Trajectories}: The observed past trajectories of all agents in the scene, \(\{\mathbf{X}_{1:T}^j\}_{j=1}^N\), where \(N\) is the total number of agents.
    \item \textit{Map Information}: A vectorized representation of the scene map, \(\mathbf{M}\), which encodes the lane geometry, such as centerlines and boundaries.
\end{enumerate}

The objective is to learn a mapping function \(f\) that predicts future trajectories, associated uncertainties, and mode probabilities:
\[
f(\mathbf{X}_{1:T}^i, \mathbf{M}; \Theta) = \{ \mathbf{\hat{Y}}_{1:T'}^{i,k}, \mathbf{\Sigma}_{1:T'}^{i,k}, p_k^i \}_{k=1}^K,
\]
where \(\Theta\) are the learnable parameters of the model.

\subsection{Model Architecture}

Our model utilizes the Hierarchical Vector Transformer (HiVT) encoder for multi-agent motion prediction~\cite{zhou2022hivt}, and extends it by incorporating an evidential framework to estimate both positional and mode uncertainty. Specifically, we model the positional uncertainty using a NIG distribution and the uncertainty in trajectory mode probabilities using a Dirichlet distribution to account for both aleatoric and epistemic uncertainty. Unlike traditional uncertainty estimation methods that require multiple forward passes, our approach directly infers uncertainty in real time with a single forward pass. Furthermore, we aggregate both point-wise positional uncertainty and mode uncertainty to provide a comprehensive agent-wise uncertainty estimate, offering a more holistic understanding of prediction confidence.\looseness=-1

\subsection{Trajectory Prediction} 

Our architecture consists of three main stages: (1) Local Context Extraction, (2) Global Interaction Modeling, and (3) Multi-modal Future Decoding (Fig.~\ref{fig:evidential}A). In the local stage, the scene is divided into agent-centric local regions, where each region extracts context features from nearby agents and lane segments using translation-invariant representations and rotation-invariant spatial learning modules. The local encoder applies self-attention mechanisms to capture fine-grained agent-agent and agent-lane interactions, followed by a temporal Transformer that learns dependencies over time. This approach allows the model to process a large number of agents efficiently while maintaining computational efficiency.

In the global interaction stage, the model aggregates information across different local regions to capture long-range dependencies and scene-level dynamics. The global Transformer module performs message passing among agent-centric embeddings while incorporating inter-agent geometric relationships. This hierarchical structure significantly reduces computational complexity compared to conventional all-to-all interaction models. Finally, the multi-modal future decoder predicts multiple possible future trajectories for all agents by predicting the parameters of a NIG mixture model to capture the inherent uncertainty in predicted position and a Dirichlet Distribution for coefficients of mixture components to capture mode probability and uncertainty at the same time.\looseness=-1

\subsection{Evidential Regression for Future Position Prediction}

Predicting the future position of an agent is inherently uncertain due to the stochastic nature of motion and perceptual noise. Instead of only predicting 2D coordinates $(x, y)$, we adopt an evidential framework to estimate both the most probable future positions and the associated uncertainty. Following the Deep Evidential Regression framework~\cite{aminiEDL}, we model positional uncertainty using the NIG distribution (Fig.~\ref{fig:evidential}B), which enables direct estimation of aleatoric and epistemic uncertainty. Instead of predicting a single deterministic value, the model outputs parameters governing a higher-order probability distribution over the unknown mean and variance.
To simplify the notations, we describe the formulation for a single position component, either $x$ or $y$. The future position component is assumed to follow a Gaussian distribution with an unknown mean $\mu$ and variance $\sigma^2$. To model this uncertainty, we place a NIG prior over these parameters:
\begin{equation}
  \begin{gathered}
    y \sim \mathcal{N}(\mu, \sigma^2) \\ 
    \mu \sim \mathcal{N}(\gamma,\sigma^{2}\nu^{-1}) ,\sigma^2 \sim \Gamma^{-1}(\alpha, \beta).
    \end{gathered}
\end{equation}

\begin{equation}
p(y | \gamma, v, \alpha, \beta) = \text{NIG}(\gamma, v, \alpha, \beta),
\end{equation}
where \( \gamma \) represents the mean of the predicted position, while \( v \) acts as an inverse variance weight. The parameters \( \alpha \) and \( \beta \) define the shape and scale of the inverse-gamma prior. The NIG hyperparameters \( m=(\gamma, v, \alpha, \beta) \) determine not only the location but also the dispersion concentrations, or uncertainty, associated with the inferred likelihood function. Therefore, the NIG distribution can be interpreted as a higher-order, evidential distribution on top of the unknown lower-order likelihood distribution from which observations are drawn.

To train our model, the loss function for the regression part consists of three terms: (1) a negative log-likelihood (NLL) loss to fit the data, (2) a regularization term to penalize incorrect certainty, and (3) a Kullback-Leibler (KL) divergence term to constrain the evidential distribution.
The primary loss function is derived from the evidential deep learning framework introduced in \cite{aminiEDL}. Given a predicted NIG distribution, the marginal likelihood of an observation \( y \) is modeled using a Student-t distribution which minimizing the NLL results in the following loss:
\begin{equation}
\begin{aligned}
\mathcal{L}_{\text{NLL}}^{reg} = & \frac{1}{2} \log \left( \frac{\pi}{v} \right) - \alpha \log(\Omega) \\
& + \left( \alpha + \frac{1}{2} \right) \log \left( (y - \gamma)^2 v + \Omega \right) \\
& + \log \left( \frac{\Gamma(\alpha)}{\Gamma(\alpha + \frac{1}{2})} \right),
\end{aligned}
\end{equation}
where \( \Omega = 2\beta(1 + v) \).

To prevent overconfidence in incorrect predictions, we add a penalty that scales with the total evidence defined as  \( 2\beta(1 + v) \), which represents the model's confidence in its prediction.
\begin{equation}
\mathcal{L}_{\text{R}}^{reg} = | y - \gamma | \cdot (2 v + \alpha).
\end{equation}

We also utilize the KL divergence term:
\begin{equation}
\mathcal{L}_{\text{KL}}^{reg} = \text{KL} \left[ p(\mu, \sigma^2 | m) \parallel p(\mu, \sigma^2 | m_0) \right],
\end{equation}
where the second term represents a zero-evidence prior distribution with minimal certainty. This term discourages the model from assigning excessive certainty to incorrect predictions. The final training loss for positional prediction incorporates these terms and will take the form:
\begin{equation}
\mathcal{L}_{reg} = \mathcal{L}_{\text{NLL}}^{reg} + \lambda_{1} \mathcal{L}_{\text{R}}^{reg} + \lambda_{2} \mathcal{L}_{\text{KL}}^{reg}.
\end{equation}

\subsection{Evidential Classification for Mode Probability Estimation}
To capture the diverse behaviors that can occur in a given scenario, trajectory prediction models generate multiple possible future trajectories and assign a probability to each mode. The uncertainty arising from the need to predict multiple plausible futures is referred to as \textit{long-term uncertainty}~\cite{shao2024uncertainty}. 
We estimate a \textit{calibrated} probability distribution over predicted modes by leveraging the classification case of the evidential framework. Instead of directly predicting a discrete probability distribution, we assume a Dirichlet prior over the mode probabilities (Fig.~\ref{fig:evidential}C). Given \( K \) possible trajectory modes, the probability distribution over modes is parameterized as
\begin{equation}
p(\mathbf{p} | \boldsymbol{\alpha}) = \text{Dir}(\mathbf{p} | \alpha_1, \alpha_2, ..., \alpha_K),
\end{equation}
where \( \mathbf{p} = (p_1, p_2, ..., p_K) \) represents the categorical probabilities assigned to each trajectory mode, and \( \boldsymbol{\alpha} = (\alpha_1, \alpha_2, ..., \alpha_K) \) are the Dirichlet concentration parameters. Higher values of \( \alpha_k \) indicate greater certainty in the corresponding mode probability.
Rather than predicting a single probability vector \( \mathbf{p} \), our model additionally estimates the evidential parameters \( \boldsymbol{\alpha} \), allowing the Dirichlet distribution to encode both the predicted probabilities and the associated uncertainty.

To train the model, we minimize a loss function consisting of two main terms: (1) a squared error term to encourage correct classification and (2) a regularization using KL divergence. The main loss function is based on the evidential framework for classification~\cite{sensoy2018evidential}, and the KL divergence term is used to penalize the deviations from a weakly informative prior, ensuring that the learned uncertainty remains well-calibrated. This results in the following classification loss, which combines a squared error term with an uncertainty regularization component:
\begin{equation}
\mathcal{L}_{squared}^{cls} = \sum_{k=1}^{K} (y_k - \frac{\alpha_k}{S})^2 + \sum_{k=1}^{K} \frac{\alpha_k (S - \alpha_k)}{S^2 (S + 1)},
\end{equation}
\begin{equation}
\mathcal{L}_{\text{KL}}^{cls} = \text{KL}\left[\text{Dir}(\mathbf{p} | \tilde{\boldsymbol{\alpha}}) \,\Big\|\, \text{Dir}(\mathbf{p} | \boldsymbol{\alpha}_0) \right],
\end{equation}
where $\boldsymbol{\alpha}_0$ is the parameters of a zero-evidence prior Dirichlet distribution and $\tilde{\boldsymbol{\alpha}}$ are Dirichlet parameters after the removal of the non-misleading evidence from predicted parameters.
Our final loss function for classification combines these terms:
\begin{equation}
\mathcal{L}_{cls} = \mathcal{L}_{squared}^{cls} + \lambda_{3} \mathcal{L}_{\text{KL}}^{cls}.
\end{equation}

The final training objective combines the evidential losses for both regression and classification, incorporating negative log-likelihood, regularization, and KL divergence, resulting in the total loss:
\begin{equation}
\mathcal{L}_{\text{total}} = \mathcal{L}_{\text{reg}} + \lambda_{4} \mathcal{L}_{\text{cls}} .
\end{equation}

This formulation ensures robust uncertainty-aware learning by jointly optimizing trajectory position uncertainty and calibrated mode probability.

\begin{table*}
\centering 
\caption{Trajectory Prediction and Uncertainty Estimation Performance Results on the Argoverse~2 dataset.} 
\label{av2}
\vspace*{-0.2cm}
\setlength\tabcolsep{3.5pt}
\begin{threeparttable}
\begin{tabular}{l|cccccccc|c|c} 
\toprule
Model &  minADE & wADE & minFDE & wFDE & MR &  minADE R-AUC & wADE R-AUC& ECE & Inference Time & Parameters \\ 
\midrule 
HIVT & 1.06  & 2.91 & 2.64 & 7.80 & 0.35  & 0.37 & 1.13 & 0.05 & $5.2\times10^{-3}$ & 666K \\ 
Error Regression & 1.06  & 2.91 & 2.64 &7.80 & 0.35 & 0.39 & \textbf{1.09}  & 0.05 & $8.6\times10^{-2}$& 670K\\ 
\midrule
MC DROPOUT & 1.06 & 3.44& 2.63 & 9.17 & 0.36  & 0.40 & 1.30& 0.06 & $5.4\times10^{-2}$& 666K\\ 

Deep Ensemble & 0.91 & 3.20& 2.15 & 8.61 & 0.31 & 0.34 & 1.19 & 0.05 & $6.2\times10^{-2}$& 3.3M  \\ 
\midrule
EVReg (Ours) & \textbf{0.83} &\textbf{2.87} & \textbf{1.78} &\textbf{7.49}& \textbf{0.29} & \textbf{0.32} & \textbf{1.09} & 0.08 & $5.6\times10^{-3}$ & 678K \\
EVReg, EVCls (Ours) & \textbf{0.83} & 2.98 & 1.80 & 7.79 & \textbf{0.29} & \textbf{0.32} & 1.16 & \textbf{0.01} & $5.6\times10^{-3}$ & 678K \\ 
\bottomrule
\end{tabular} 
\end{threeparttable}
\vspace{-0.3cm}
\end{table*}

\subsection{Uncertainty Aggregator Module}
Our framework estimates the uncertainty at per-axis level in regression, and it needs to be aggregated to obtain a meaningful uncertainty measure of an agent's prediction. In~\cite{shao2024uncertainty}, a comprehensive evaluation was conducted on how to aggregate risk from point-wise uncertainty to an agent level.
Different formulations of agent-level uncertainty can be considered depending on which metric is prioritized for the task. For instance, if we focus solely on minimum Final Displacement Error (minFDE), we can derive uncertainty from the trajectory that is closest to the ground truth and consider only the point-wise uncertainty at the final position, but the most complete formulation accounts for all predicted trajectories and all time steps. To obtain the uncertainty of an agent, we aggregate uncertainty from the per-axis level up to the full trajectory and across multiple modes. This process involves the following steps:
\begin{itemize}
    \item Each axis \( x \) and \( y \) has two sources of uncertainty: \textit{epistemic} \( U_{\text{epistemic}} \) and \textit{aleatoric} \( U_{\text{aleatoric}} \). The total uncertainty for each axis is defined as:
    \begin{align}
        U_{\text{aleatoric}} &= \frac{\beta}{\alpha-1}, U_{\text{epistemic}} = \frac{\beta}{(\alpha-1)\nu}, \\
        U_{\text{total}, x} &= U_{\text{epistemic}, x} + U_{\text{aleatoric}, x}, \\
        U_{\text{total}, y} &= U_{\text{epistemic}, y} + U_{\text{aleatoric}, y}.
    \end{align}
    This step ensures that both the model’s confidence in its predictions (epistemic) and the inherent noise in the data (aleatoric) contribute to the final uncertainty measure.

    \item  Since trajectory predictions are in a 2D space, we compute a scalar point-wise uncertainty by summing the uncertainties in both axes:
    \begin{equation}
        U_{\text{point}} = U_{\text{total}, x} + U_{\text{total}, y}.
    \end{equation}
    This sum reflects the total spatial uncertainty at each predicted point, ensuring that both horizontal and vertical deviations are accounted for.

    \item To obtain an uncertainty measure for an entire predicted trajectory, we take the average uncertainty across all time steps in each trajectory:
    \begin{equation}
        U_{\text{traj}, k} = \frac{1}{T'} \sum_{t'=1}^{T'} U_{\text{point}, t'}^{(k)}.
    \end{equation}

    Here, \( T' \) represents the total number of future timesteps, and \( U_{\text{point}, t'}^{(k)} \) is the uncertainty at timestep \( t' \) for trajectory mode \( k \) which ensures that trajectories with more uncertain steps contribute more to the overall measure. 
    
    \item Instead of weighting each trajectory’s uncertainty by its predicted probability \( p_k \), we can weight it based on its classification uncertainty. The classification uncertainty for K mode classification is defined as:
    \begin{equation}
        U_{\text{cls}} = \frac{K}{S} = \frac{K}{\sum_{j=1}^{K} \alpha_j}.
    \end{equation}
    A lower total evidence \( S \) results in higher uncertainty, as the model lacks confidence in selecting a mode. This classification uncertainty is used to scale the total trajectory uncertainty across all modes. The final agent-level uncertainty is computed as:
    \begin{equation}
    U_{\text{agent}} = \frac{1}{K} \sum_{k=1}^{K} U_{\text{cls}} \cdot U_{\text{traj}, k}.
    \end{equation}
\noindent This formulation ensures that both trajectory uncertainty and classification uncertainty are integrated into a single uncertainty measure.
    
\end{itemize}

\section{Experiments}

In this section, we describe the experiments that we perform, including the evaluation framework and results.

\subsection{Experimental Setup}

We evaluate our method on the Argoverse~2 (Av2)\cite{wilsonArgoverse} dataset and validate its generalizability on Argoverse~1 (Av1)\cite{chang2019argoverse} dataset. Av2 contains 763 hours of driving data across 250,000 scenarios (\SI{11}{\second} each), while Av1 includes 320 hours with 323,577 segments (\SI{5}{\second} each). Both datasets are sampled at 10 Hz, but Av1 provides \SI{2}{\second} of past observations, whereas Av2 uses \SI{5}{\second} as input, with the remaining duration as ground truth.
Training was conducted on 8 NVIDIA V100 GPUs with a learning rate of $9 \times 10^{-4}$, weight decay of $1 \times 10^{-4}$, and batch size of $16$. Av1 models were trained for up to $64$ epochs with early stopping (patience of 10 epochs, minimum improvement threshold of $1 \times 10^{-3}$). Given Av2's larger size, training was extended to 70 epochs while keeping the early stopping criteria unchanged.

\subsection{Evaluation Metrics}
To assess trajectory prediction accuracy, we rely on the following metrics:
\textit{minADE} (average displacement error of the closest trajectory),  
\textit{minFDE} (final-step displacement error of the closest trajectory),  
\textit{wADE} (ADE weighted by mode probability), and  
\textit{wFDE} (FDE weighted by mode probability).  
For uncertainty calibration, we use:  
\textit{R-AUC}~\cite{malinin2021shifts} (correlation between uncertainty and error via progressive sample rejection) and  
\textit{ECE}~\cite{ece} (alignment of predicted mode probabilities with correctness likelihoods).  
Additionally, we report \textit{Inference time} and \textit{Number of parameters} to assess computational efficiency.

\subsection{Baselines for Uncertainty Estimation}
We compare with the following uncertainty estimation approaches commonly used in trajectory prediction:
\paragraph{HiVT}
The model estimates aleatoric positional uncertainty by fitting a Laplace distribution to each 2D trajectory point.

\paragraph{Error Regression}
Following \cite{wiederer2023joint}, we train an MLP to predict the error of a trained trajectory prediction model. A new dataset is generated where the input consists of latent features, and the output is the trajectory prediction error.

\paragraph{MC Dropout}
In this method, dropout layers remain active at inference time, and N stochastic forward passes are performed. The variance of the predicted trajectories is then used as an uncertainty estimate. 

\paragraph{Deep Ensemble}
Instead of relying on dropout, Deep Ensemble trains N independent models with different random initializations. At inference time, predictions from all models are aggregated, and the variance is treated as the uncertainty

Since MC Dropout and Deep Ensemble generate 30 (modes times inference passes) trajectory candidates (by leveraging multiple stochastic passes), they have an inherent advantage in minADE and minFDE and consequently on minADE R-AUC. These metrics select the closest trajectory to the ground truth, and having a larger candidate pool increases the likelihood of a better match. Hence, we applied a k-means clustering on the agents' predicted trajectories and used the cluster average as the predicted mode to make the comparison meaningful.

\begin{table*}
\centering
\caption{Trajectory prediction and uncertainty estimation performance comparison on Argoverse~1.1.}
\label{tab:av1}
\vspace*{-0.2cm}
\begin{tabular}{l|cccccccc} 
\toprule
Model & minADE & wADE & minFDE & wFDE & MR & minADE R-AUC & wADE R-AUC  & ECE \\ 
\midrule
HIVT & 0.62  & \textbf{1.85} & 1.07 & \textbf{4.10} & 0.11  & 0.25 & 0.74 &  0.04\\ 
Error Regression & 0.62  & \textbf{1.85} & 1.07 & \textbf{4.10} & 0.11  & 0.24 & \textbf{0.70} & 0.04\\
\midrule
MC DROPOUT & 0.63 & 2.17 & 1.10 & 4.84 & 0.10 & 0.24 & 0.83 & 0.05\\ 
Deep Ensemble & 0.61 & 2.09 & 1.04 & 4.64 & 0.10 & \textbf{0.23} & 0.78 & 0.04\\
\midrule
Ours EvReg& 0.61 & 1.88 & \textbf{1.01} & 4.12 & \textbf{0.09} & 0.24 & 0.74 & 0.05\\ 
Ours EvReg,EvCls& \textbf{0.60} & 1.94 & \textbf{1.01} & 4.33 & 0.10 & \textbf{0.23} & 0.75 & \textbf{0.00}\\ 
\bottomrule
\end{tabular} 
\end{table*}

\begin{table*}
\centering 
\caption{Importance sampling experiment results.} 
\vspace*{-0.2cm}
\label{IS}
\begin{tabular}{l|ccccccc} 
\midrule 
Model & minADE & wADE & minFDE & wFDE & MR  & minADE R-AUC & wADE R-AUC  \\ 
\midrule
Base & 0.86 & 2.93 & 1.87 & 7.68 & 0.31 & 0.33& 1.14 \\ 
\midrule
Random 75\% & 0.85 & 2.92 & 1.84 & 7.64 & 0.30 & 0.32 & 1.13 \\ 
Error 75\% & \textbf{0.81} & 2.80 & 1.74 & 7.38 & 0.29 & \textbf{0.31} & 1.08 \\ 
Uncertainty 75\% & \textbf{0.81} & \textbf{2.77} & \textbf{1.73} & \textbf{7.30} & \textbf{0.28}  & \textbf{0.31} & \textbf{1.06}  \\ 
\midrule
Full training & 0.83 & 2.87 & 1.78 & 7.49 & 0.29  & 0.32 & 1.09  \\
\hline 
\end{tabular}
\vspace{-0.4cm}
\end{table*}

\subsection{Quantitative Results}
Tab.~\ref{av2} and Tab.~\ref{tab:av1} present a quantitative comparison of our model against several baselines on the Av2 and Av1 datasets, respectively. Our evidential regression (EvReg) model achieves the best performance across several key metrics, particularly in minADE of 0.83 and minFDE of 1.78 on Av2 and respective values of 0.60 and 1.01 on Av1. Compared to traditional uncertainty estimation approaches, our model provides accurate trajectory predictions while maintaining significantly lower inference times. Specifically, our method runs in \(5.6 \times 10^{-3}\) sec per prediction}, making it almost nine times faster than sampling-based methods. This highlights the efficiency of our evidential framework, which estimates uncertainty in a single forward pass instead of requiring multiple stochastic samples. One trade-off of our approach is a slight degradation in wADE and wFDE scores compared to other baselines. However, this is counterbalanced by a significant improvement in probability calibration, as evidenced by the ECE dropping to 0.01. This suggests that while our method may assign lower probabilities to the closest trajectory compared to overconfident baselines, the predicted probabilities better reflect the true uncertainty.

Moreover, methods such as Error Regression are explicitly trained to predict model error, therefore, they have an inherent advantage in R-AUC, achieving minADE R-AUC of 0.3 and wADE R-AUC of 1.09 on Av2 and respective values of 0.24 and 0.70 on Av1. However, a key limitation of error regression models is their reliance on a separate uncertainty estimation module rather than being integrated within an end-to-end learning framework. On top of that, the error that is used to train them plays an important role in how they perform on the R-AUC metric as we can see they perform better on predicting error with wADE compared to minADE. Overall, the results demonstrate that our evidential framework provides a balanced approach, achieving high accuracy, efficient uncertainty estimation, and better-calibrated probability while maintaining real-time inference capabilities.

\begin{figure}
\centering
\includegraphics[width=\columnwidth]{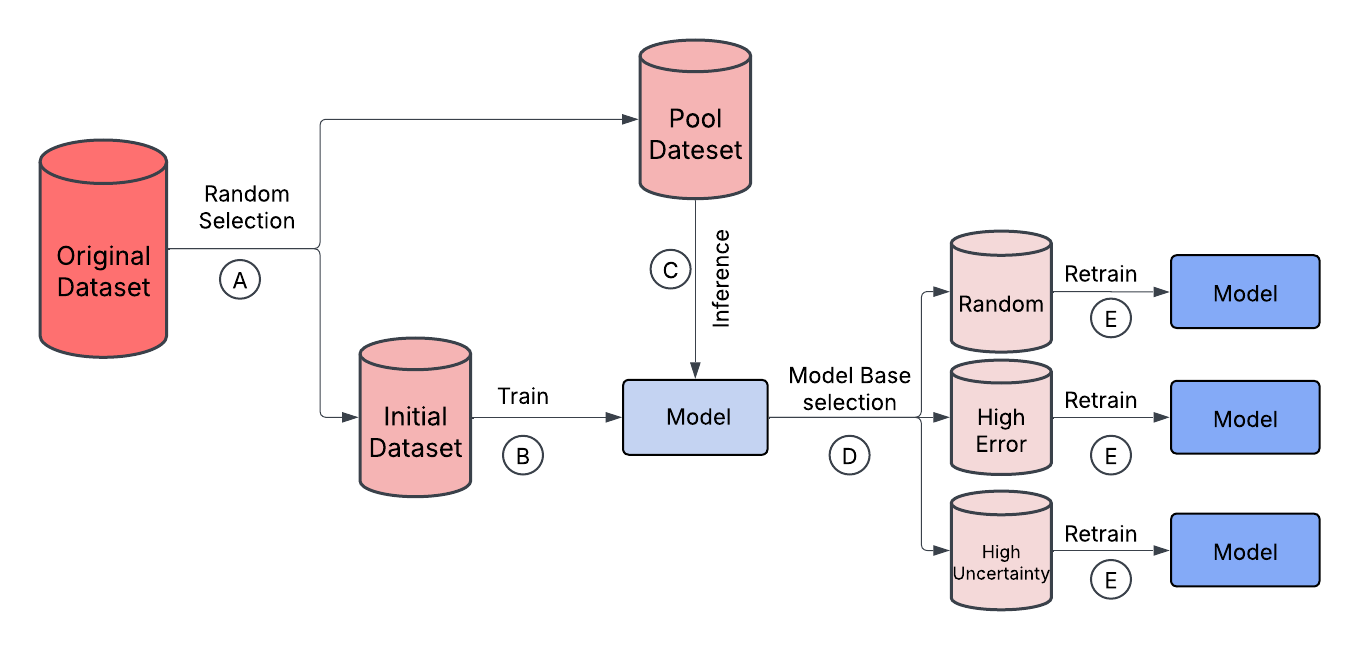}
\caption{Our Importance sampling framework. A model is trained on an initial randomly selected dataset, then the trained model is used to select new samples from remaining data points to retrain.}
\label{fig:importance_sampling}
\vspace{-0.4cm}
\end{figure}

\subsection{Importance Sampling Experiment}

To investigate whether the predicted uncertainty can guide data selection and improve the efficiency of training, we conduct an importance sampling experiment (Fig.~\ref{fig:importance_sampling}):
\begin{enumerate}
    \item Train the model on 50\% of the Argoverse~2 dataset.
    \item From the remaining 50\%, select half (25\% of total data) using three different criteria:
        \begin{enumerate}
            \item \textit{Random selection}, used as a baseline.
            \item \textit{Error-based selection}, where samples with the highest minADE error are chosen. This method relies on the ground truth, making it impractical for real-time applications.
            \item \textit{Uncertainty-based selection}, where samples with the highest predicted uncertainty are selected.
        \end{enumerate}
    \item Train the model on the selected 75\% subset for an additional 40 epochs, applying early stopping.
\end{enumerate}

The goal of this experiment is to determine whether uncertainty estimates can identify high-value training samples that contribute to greater model improvement. With autonomous driving datasets growing exponentially, storing and processing all collected data is increasingly challenging. By selectively training on the most informative samples, we aim to reduce data requirements while maintaining strong predictive performance. The results from this experiment are presented in Tab.~\ref{IS}, where our model, starting from the same checkpoint, demonstrates improved performance across almost all metrics when using uncertainty-based selection. Notably, with just 75\% of the data, our model outperforms the fully trained model on 100\% of the data, highlighting the advantage of prioritizing non-trivial scenarios. Unlike error-based selection, which requires ground truth information for the remaining 50\% of the dataset, our uncertainty-based approach operates without access to ground truth, making it a more practical and scalable solution.

\subsection{Qualitative Results}

\begin{figure}
\centering
\includegraphics[width=\columnwidth]{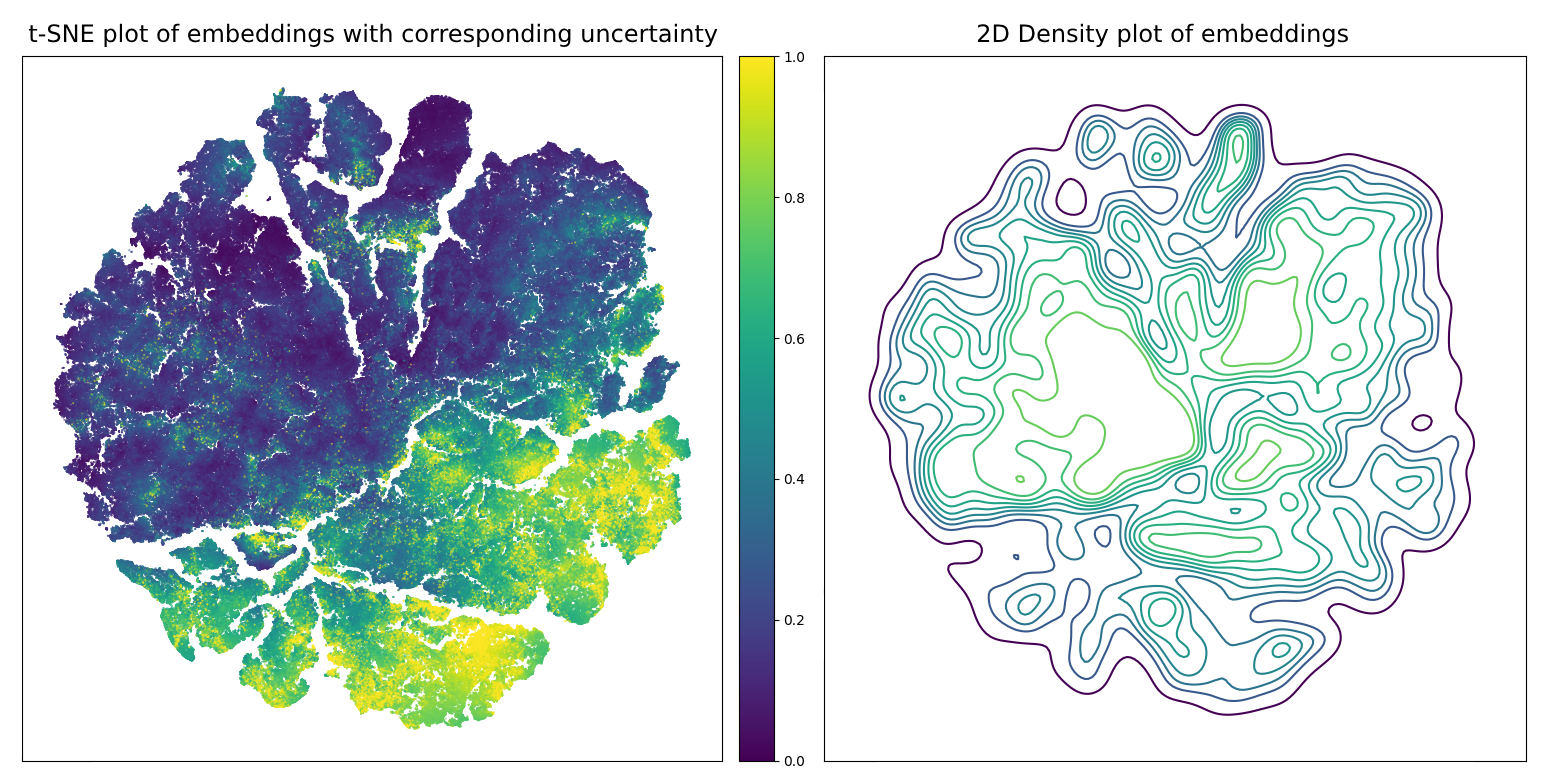}
\caption{t-SNE plot of Argoverse~2 validation set. All the embeddings have been mapped to 2D and color-coded with uncertainty on the left and the density of samples on the right.}
\label{fig:combined_plots_with_kde}
\vspace{-0.4cm}
\end{figure}

We investigate the relationship between learned latent space and predicted uncertainty using t-SNE visualizations. The results in Fig.~\ref{fig:combined_plots_with_kde} show an inverse correlation between sample density and uncertainty, where denser regions exhibit lower uncertainty, while sparse regions correspond to higher uncertainty.
This suggests that our evidential model assigns higher confidence to predictions in well-represented regions of the latent space, reinforcing the idea that the model captures uncertainty in a data-driven manner.

\begin{figure}
\centering
\includegraphics[width=\columnwidth]{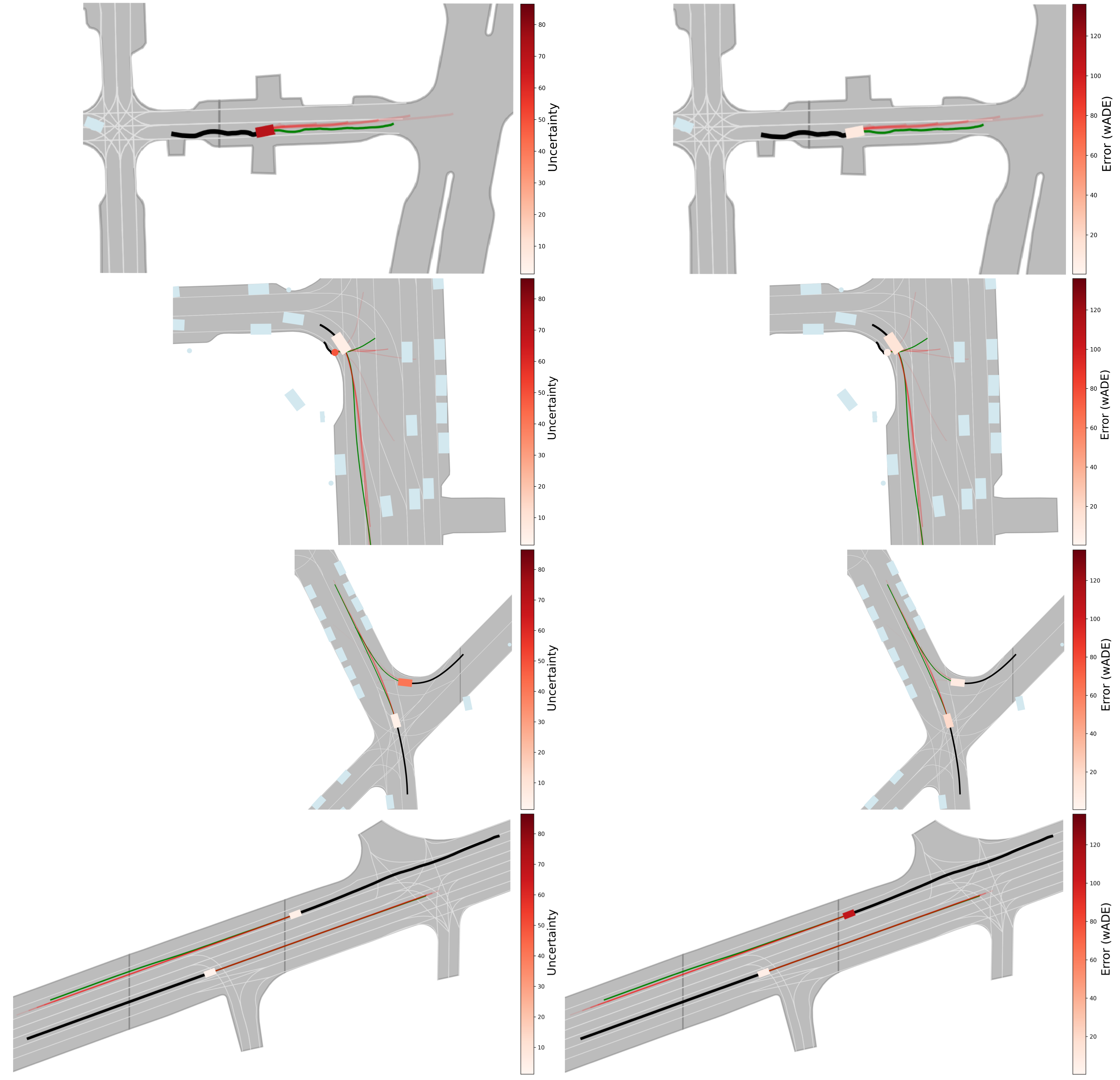}
\caption{Trajectory prediction results on Argoverse~2 validation set. Agent history is in black, ground truth in green, and predicted trajectories in red with opacity reflecting mode probability. Left: Agent colors encode uncertainty. Right: Colors encode prediction error.}
\label{fig:traj}
\vspace{-0.4cm}
\end{figure}

Fig.~\ref{fig:traj} presents qualitative results on the Argoverse~2 validation dataset, where the left column visualizes uncertainty-based encoding, while the right column encodes prediction error (wADE). The first three rows show examples selected based on high uncertainty, whereas the last row is chosen based on high prediction error. A key observation is that higher uncertainty is typically associated with complex or rare driving scenarios. For instance, in the second row, where a pedestrian is in close proximity to a vehicle, the model exhibits increased uncertainty, likely due to the ambiguity in predicting whether the vehicle will yield or continue moving. Similarly, in the first and third rows, we observe jerky driving behaviors and scenarios where an agent's decision-making depends on other vehicles’ actions. These cases introduce unpredictability, leading to higher model uncertainty and reflecting the model's awareness of its limitations in less frequently observed situations.

In contrast, the last row which was selected for high prediction error, primarily features high-speed driving scenarios. While the trajectories appear simple (mostly straight lane driving), the error accumulates over longer distances. Even small deviations in prediction can lead to significant displacement errors, particularly in minFDE and wADE metrics. This suggests that uncertainty highlights decision-dependent situations, whereas high prediction error is more linked to fast, linear motion, where minor inaccuracies compound over time. These visualizations emphasize the importance of distinguishing predictive uncertainty from prediction error, as they reveal different aspects of model reliability and improvement areas.

\section{Conclusion}

In this work, we presented a novel evidential deep learning framework for real-time trajectory prediction and uncertainty estimation. Our framework outperforms baseline methods in predicting calibrated probabilities, trajectory accuracy, and agent-wise uncertainty. Qualitative results further revealed that high-uncertainty agents tend to correspond to more complex scenarios, distinguishing them from trivial cases. Additionally, we effectively leveraged predicted uncertainty as a signal for data importance, enabling more efficient training. Notably, by training on only 75\% of the data, our model surpassed the performance of training on the full dataset, highlighting the benefits of uncertainty-aware learning.

\bibliographystyle{IEEEtran}
\bibliography{main}

\end{document}